\documentclass[conference]{IEEEtran}
\IEEEoverridecommandlockouts
\usepackage{cite}
\usepackage{amsmath,amssymb,amsfonts}
\usepackage{algorithmic}
\usepackage{graphicx}
\usepackage{textcomp}
\usepackage{xcolor}
\usepackage{hyperref}
\usepackage{multirow}
\usepackage{array}
\def\BibTeX{{\rm B\kern-.05em{\sc i\kern-.025em b}\kern-.08em
    T\kern-.1667em\lower.7ex\hbox{E}\kern-.125emX}}

\usepackage{listings}
\usepackage{color}
\usepackage{tabularray}

\definecolor{dkgreen}{rgb}{0,0.6,0}
\definecolor{gray}{rgb}{0.5,0.5,0.5}
\definecolor{mauve}{rgb}{0.58,0,0.82}

\begin{document}

\title{QASiNa: Religious Domain Question Answering using Sirah Nabawiyah\\

}

\author{\IEEEauthorblockN{
Muhammad Razif Rizqullah\textsuperscript{\textbf{1}}, 
Ayu Purwarianti\textsuperscript{\textbf{1}},
Alham Fikri Aji\textsuperscript{\textbf{2}}
}
\IEEEauthorblockA{
\textsuperscript{1}School of Electrical Engineering and Informatics \\
Bandung Institute of Technology. Bandung, Indonesia \\
razifrizqullah@gmail.com, ayu@stei.itb.ac.id}
\IEEEauthorblockA{
\textsuperscript{2}Mohamed bin Zayed University of Artificial Intelligence\\
Abu Dhabi, UAE \\
alham.fikri@mbzuai.ac.ae}
}

\maketitle

\begin{abstract}
Nowadays, Question Answering (QA) tasks receive significant research focus, particularly with the development of Large Language Model (LLM) such as Chat GPT \cite{chatgpt}. LLM can be applied to various domains, but it contradicts the principles of information transmission when applied to the Islamic domain. In Islam we strictly regulates the sources of information and who can give interpretations or tafseer for that sources \cite{solahudin2016pendekatan}. The approach used by LLM to generate answers based on its own interpretation is similar to the concept of tafseer, LLM is neither an Islamic expert nor a human which is not permitted in Islam. Indonesia is the country with the largest Islamic believer population in the world \cite{annur2023ini}. With the high influence of LLM, we need to make evaluation of LLM in religious domain. Currently, there is only few religious QA dataset available and none of them using Sirah Nabawiyah especially in Indonesian Language. In this paper, we propose the Question Answering Sirah Nabawiyah (QASiNa) dataset, a novel dataset compiled from Sirah Nabawiyah literatures in Indonesian language. We demonstrate our dataset by using mBERT \cite{devlin2018bert}, XLM-R \cite{conneau2019unsupervised}, and IndoBERT \cite{koto2020indolem} which fine-tuned with Indonesian translation of SQuAD v2.0 \cite{muis2020sequence}. XLM-R model returned the best performance on QASiNa with EM of 61.20, F1-Score of 75.94, and Substring Match of 70.00. We compare XLM-R performance with Chat GPT-3.5 and GPT-4 \cite{chatgpt}. Both Chat GPT version returned lower EM and F1-Score with higher Substring Match, the gap of EM and Substring Match get wider in GPT-4. The experiment indicate that Chat GPT tends to give excessive interpretations as evidenced by its higher Substring Match scores compared to EM and F1-Score, even after providing instruction and context. This concludes Chat GPT is unsuitable for question answering task in religious domain especially for Islamic religion.
\end{abstract}


\section{Introduction}

Question answering is a task that closely aligns with everyday human behavior based on the theory of mind \cite{nematzadeh2018evaluating}. In daily human interactions, discussions are conducted to exchange information, starting with one person providing a statement or question, followed by response or answer. Currently, question answering methods have significantly advanced, including rule-based approaches, extractive language models utilizing reading comprehension, and Large Language Model (LLM) with generative approach. These methods are commonly employed for general question answering problem, but there is limitations when applied to specific domains such as the religious domain, especially Islamic religion.

\begin{figure}
  \centering
  \includegraphics[width=0.4\textwidth]{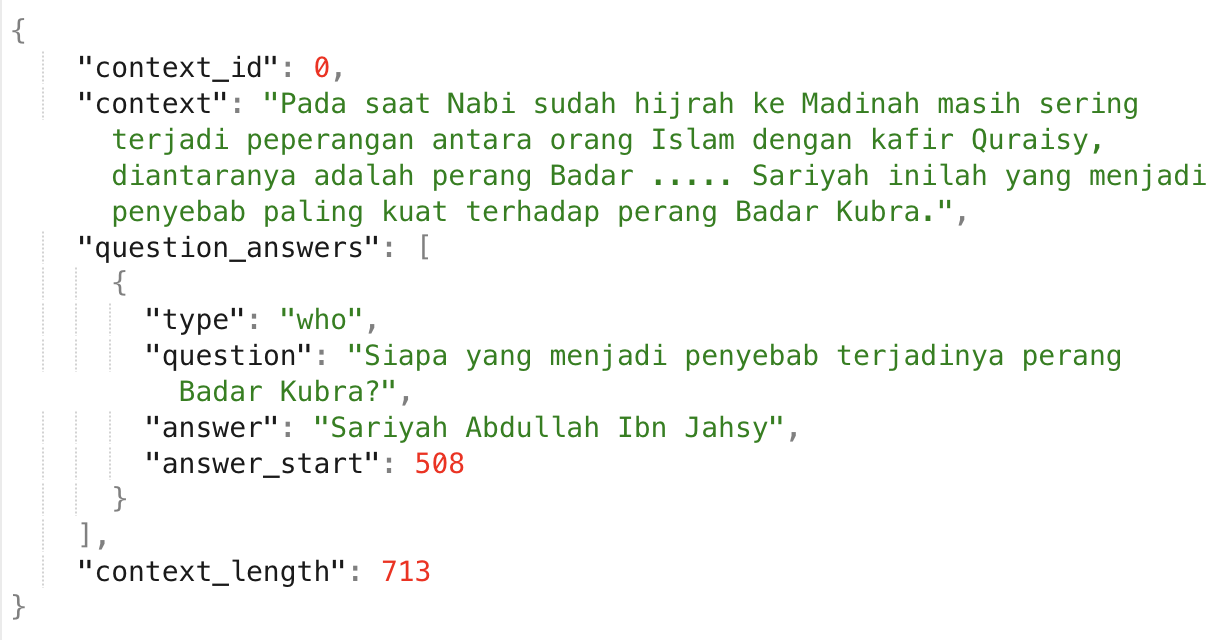}
  \caption{Example of context, question, and answer in QASiNa dataset}
  \label{fig:data-example}
\end{figure}

Indonesia is the country with the largest Islamic believer (moslem) population in the world. According to Annur \cite{annur2023ini} on the "katadata" website, the moslem population is 237.558 million people, 86.7\% from total population, based on a survey conducted by The Royal Islamic Strategic Studies Centre (RISSC) in 2023. The primary direct references for moslem are the Holy Qur'an and Sunnah from Book of Hadith \cite{jaya2019qur}. In addition to these two primary sources, Sirah Nabawiyah (Prophetic Biography) serves as another important reference because it contains message, vision, mission, and historically activities to support both primary references \cite{al2012sirah}. Sirah Nabawiyah contains comprehensive explanation about history of Islam from before the birth of Prophet Muhammad until his passing and the continuation of Islamic preaching.

According to Solahudin \cite{solahudin2016pendekatan}, interpretation or reasoning methods in Islamic information sources can be divided into two types, contextual and textual reasoning. Contextual reasoning requires extensive and comprehensive knowledge which requires learning process under supervision of experts, examination, and scholarly validation. On the other hand, textual reasoning is a method that involves extracting information directly from the context without providing interpretation. 

With current huge influence of LLM in daily basis, a country with the largest moslem population will face a problem because LLM works by giving its own interpretation using next token prediction. Islamic resource interpretation for public audience is called tafseer. Based on tafseer rule in Islam, tafseer can only be given by a person who specialized at it \cite{mutalib2019scientific}. Islamic society need a way to evaluate how far the interpretation given by LLM. In this paper, we conduct a research to measure textual reasoning performance of LLM Chat GPT \cite{chatgpt} compared to extractive languages models to answer religious QA with given context.

Previous research of question answering task in Islamic domain has utilized information from the Holy Qur'an \cite{abdelnasser2014bayan} \cite{malhas2022qur} \cite{malhas2020ayatec} \cite{alqahtani2018annotated} \cite{gusmita2014rule} \cite{sukmana2016semantically} \cite{putra2016semantic}, Book of Hadith \cite{abdi2020question} \cite{neamah2017question}, Islamic fatwa websites \cite{munshi2021towards} \cite{mohammed2022english}, with only a few addressing Islamic history \cite{naf2016eliminating}. Most of them are in Arabic language and none have used literature of the Indonesian Sirah Nabawiyah. 

In this research, we propose a new dataset to undertake the limitation of resources, especially in Indonesian language and the usage of Sirah Nabawiyah as source of information. There are three main outcomes of this research: 1) New dataset consisting of 500 question-answer pairs from 66 different contexts about Sirah Nabawiyah; 2) Evaluation of the extractive question answering using language models with transfer learning; and 3) Evaluate the performance of Chat GPT when answering extractive question in religious domain.

\section{Related Works}

This section presents overview of question answering methods, datasets, related question answering in religious domain, and transfer learning technique. 

\subsection{Question Answering}

The processing of information in computers until the stage where computers can perform tasks like humans is one of the goals for successful artificial intelligence \cite{nematzadeh2018evaluating}. One of the common information exchange processes performed by humans is discussion, which is similar to question answering task. There are several methods that can be used for the question answering task, namely rule-based, extractive, and generative.

Based on studies of rule-based question answering \cite{abdelnasser2014bayan} \cite{alqahtani2018annotated} \cite{abdi2020question} \cite{neamah2017question}, the working principle of rule-based approach revolves around obtaining answers using predefined patterns. Extractive method has been conducted in several studies \cite{munshi2021towards} \cite{malhas2022qur}. The extractive method utilizes context or passage, questions, and answers, where the answers are obtained from the context in the form of answer spans. The generative method which used by LLM \cite{chatgpt}, differs from extractive methods because LLM can answer questions with or without context. The generative model provides answers based on the information it has been trained on. In this research we evaluate the performance of extractive question answering using IndoBERT \cite{koto2020indolem}, XLM-R \cite{conneau2019unsupervised}, mBERT \cite{devlin2018bert} and generative model Chat GPT \cite{chatgpt}.

\subsection{Religious Domain Question Answering}

Research on the religious domain is a sensitive matter because all of the informations must align with approved sources by all followers \cite{vieten2016competencies}. Therefore, conducting research on the use of AI in the field of religion, especially Islam, is both intriguing and challenging. Several previous studies have been conducted on question answering in the religious domain.

Holy Qur'an used by Abdelnasser et al. \cite{abdelnasser2014bayan} to develop a QA dataset called Al-Bayan in Arabic language. Malhas et al. \cite{malhas2022qur} examined the Qur'anic Reading Comprehension Dataset (QRCD) from the Qur'an in Arabic. This study \cite{malhas2022qur} was an extension of a previous work by Malhas and Elsayed \cite{malhas2020ayatec} on Ayatec, a reusable verse-based test QA collection of Qur'an in Arabic. Alqahtani and Atwell \cite{alqahtani2018annotated} researched an ontology dataset called the Arabic Quranic Question and Answer Corpus (AQQAC). Sunnah from Book of Hadith used by Abdi et al. \cite{abdi2020question} to explore the use of Sahih al-Bukhari in Arabic for QA using semantic, word-order, and sentence similarity. Neamah and Saad \cite{neamah2017question} investigated the usage of Sahih al-Bukhari in English for QA, employing cosine similarity, longest common subsequence, and support vector machine (SVM). Web resources used by Munshi et al. \cite{munshi2021towards} for QA of fatwa system, using question-answer data from Arabic fatwa websites. Mohammed et al. \cite{mohammed2022english} investigated the English Islamic Article Dataset (EIAD) for Islamic chatbot.

Indonesian translation of Holy Qur'an used by Gusmita et al. \cite{gusmita2014rule} for rule-based question answering for the translation of the Holy Qur'an, which was later expanded by Sukmana et al. \cite{sukmana2016semantically} for the semantic annotated corpus method. The resulting corpus was then utilized by Putra et al. \cite{putra2016semantic} for semantic question answering using the inverted index method to search for answer candidates. Named entity recognition and feature extraction were employed to obtain the best verse and answer. Historical literature used by Naf'an et al. \cite{naf2016eliminating} to examine unanswerable question answering using Khulafaa Al-Rashidin History in the Indonesian language, employing search methods and answer candidate ranking. 

Other than Islamic domain, Zhao and Liu\cite{zhao2018finding} conducted research about Bible and created a BibleQA based on Bible trivia questions. Only a small number of QA research on religious domain exists other than Islamic, because Islam has strict rules regarding information transmission and interpretation, an interesting topic to research.

Previous studies in Islamic religion have focused primarily on the Holy Qur'an, Hadith literature, and website information, with a small number of research conducted on utilizing historical material or Sirah Nabawiyah. To tackle the data limitation, in this research we utilize literature from the Indonesian language of Sirah Nabawiyah to make a new dataset.

\subsection{Transfer Learning for Low Resources}

Developing a model from scratch using data in a specific domain requires significant time and resources. Therefore, transfer learning is employed \cite{torrey2010transfer}. Transfer learning is a machine learning technique where a model is trained on a larger dataset from a more general domain, and then the model is evaluated on a more specific domain. Previous research on question answering has utilized transfer learning methods, with the use of language models such as BERT \cite{duan2022enhancement}, RoBERTa \cite{bachina2021ensemble}, and IndoBERT \cite{rahajeng2021indonesian}. The results have shown that transfer learning can effectively answer questions in specific domains.

In this study, we propose a new dataset focused on information from the Sirah Nabawiyah which has low resources of data. Hence, we employ transfer learning methods for model training. We utilize mBERT \cite{devlin2018bert}, XLM-R \cite{conneau2019unsupervised}, and IndoBERT \cite{koto2020indolem} with transfer learning from Indonesian translation of SQuAD v2.0 (SQuAD-ID) \cite{muis2020sequence}.

\section{Question Answering Sirah Nabawiyah Dataset}

This section presents methods to build the Question Answering Sirah Nabawiyah (QASiNa) dataset which is available to be accessed from public repository\footnote{https://github.com/rizquuula/QASiNa}. The process starts from data acquisition, context retrieval, question and answer generation, and dataset validation.

\subsection{Data Acquisition}\label{DataAcquisition}

We select the data sources by searching for research literature on the Sirah Nabawiyah from various campus repositories, primarily from Islamic universities in Indonesia. The chosen literature meets specific criteria: 1) It is a valid and reliable source, proven by its publication through academic mechanisms, and 2) It is publicly accessible, allowing us to share its content in the form of the QASiNa dataset. In this phase, we obtained 9 literature sources, as detailed in Table \ref{table:literature-sources}.

\begin{table}
\centering
\caption{Sirah Nabawiyah Literature Sources}
\label{table:literature-sources}
\begin{tabular}{p{0.3cm} p{1.5cm} p{5.5cm}}
\hline
\textbf{No} & \textbf{Author} & \textbf{Title} \\
\hline
\hline
1 & Bastari \cite{bastari2013kontemplasi} & Kontemplasi Politik (Belajar Dari Kisah Perang Badar Menurut Sirah Ibnu Hisyam Dan Al-thabari) \newline \scriptsize \textit{(Political Contemplation (Learning from the Story of the Battle of Badr According to the Sirah of Ibn Hisham and Al-Thabari))}\\
2 & Zakaria \cite{zakaria2019isra} & Isra Mi’raj Sebagai Perjalanan Religi: Studi Analisis Peristiwa Isra Mi’raj Nabi Muhammad Menurut Al Qur’an Dan Hadits \newline \scriptsize \textit{(Isra Mi'raj as a Religious Journey: An Analytical Study of the Isra Mi'raj Event of Prophet Muhammad According to the Qur'an and Hadiths)} \\
3 & Salahuddin \cite{salahuddin2017isra} & Isra’Mi’raj studi analisis sejarah dalam pendidikan islam \newline \scriptsize \textit{(Isra' Mi'raj is a study of historical analysis in Islamic education)} \\
4 & Mubarok \cite{mubarok2020sejarah} & Sejarah Sosial-Politik Arab: Dari Hegemoni Romawi-Persia Hingga Kebangkitan Arab Islam \newline \scriptsize \textit{(Arab Social-Political History: From Roman-Persian Hegemony to the Rise of Arab Islam)} \\
5 & Izzani and Rubini \cite{izzani2021pendidikan} & Pendidikan Karakter dalam Buku Sirah Nabawiyah Karya Syaikh Shafiyyurrahman al-Mubarakfuri \newline \scriptsize \textit{(Character Education in the Book of the Prophet's Biography by Sheikh Shafiyyurrahman al-Mubarakfuri)} \\
6 & Hasbillah \cite{hasbillah2012sirah} & Sirah Nabawiyah dan Demitologisasi Kehidupan Nabi \newline \scriptsize \textit{(Prophetic Biography and Demythologization of the Prophet's Life)} \\
7 & Fuad \cite{fuad2014sejarah} & Sejarah peradaban Islam \newline \scriptsize \textit{(History of Islamic Civilization)} \\
8 & Thabrani \cite{thabrani2014tata} & Tata kelola pemerintahan negara madinah pada masa nabi Muhammad saw \newline \scriptsize \textit{(The governance system of the Madinah state during the time of Prophet Muhammad (peace be upon him).)} \\
9 & Sairazi \cite{sairazi2019kondisi} & Kondisi Geografis, Sosial Politik, dan Hukum Di Makkah dan Madinah Pada Masa Awal Islam \newline \scriptsize \textit{(Geographical, Social, Political, and Legal Conditions in Mecca and Medina during the Early Period of Islam)} \\
\hline
\end{tabular}
\end{table}

\subsection{Context Retrieval}\label{ContextRet}

We select contexts from each literature source and each context constitutes a complete story containing various facts within it and manually choose these contexts. Each context consists of multiple sentences and paragraphs, with the length ranging from 500 to 2000 characters. At this stage, we have obtained 66 contexts that will proceed to the question and answer generation phase. We carefully maintain a wide diversity of topics covered. We start by providing historical context, beginning with the history of the Arab region, the surrounding kingdoms, and their culture before the birth of Prophet Muhammad SAW. We also include historical context covering Muhammad SAW's early years, becoming a Prophet and building a country, and the period after the Prophet's passing.

\subsection{Question and Answer Generation}\label{QAGen}

To make the process of initial question and answer pair generation faster, we use machine assistance rather than creating question and answer pairs manually. Similar approach has been conducted in the past for the creation of dataset Indo-SQuAD v2.0 \cite{muis2020sequence} using machine translation and IDK-MRC \cite{putri2022idk} with the aid of a question generation model. In this study, we utilized a generative model Chat GPT-3.5 to generate initial question and answer pairs. We limited the generated results to 500 question and answer pairs. These question and answer pairs are not considered valid data as they were generated by a machine. Therefore, we proceeded to manually validate the data with the assistance of domain experts.

\subsection{Dataset Validation}\label{DatasetVal}

We employ domain experts to validate the context, question, and answer pairs generated by the machine. To ensure the quality of the validation process, each expert must meet the following criteria: 1) Expert is islamic believer; 2) Possess proficiency in the Indonesian language with good grammar and vocabulary; 3) Have done intensive study of Islamic knowledge, either formally and informally for a minimum of two years; and 4) Demonstrate a comprehensive understanding of context in Sirah Nabawiyah. 

\begin{figure}
  \centering
  \includegraphics[width=0.3\textwidth]{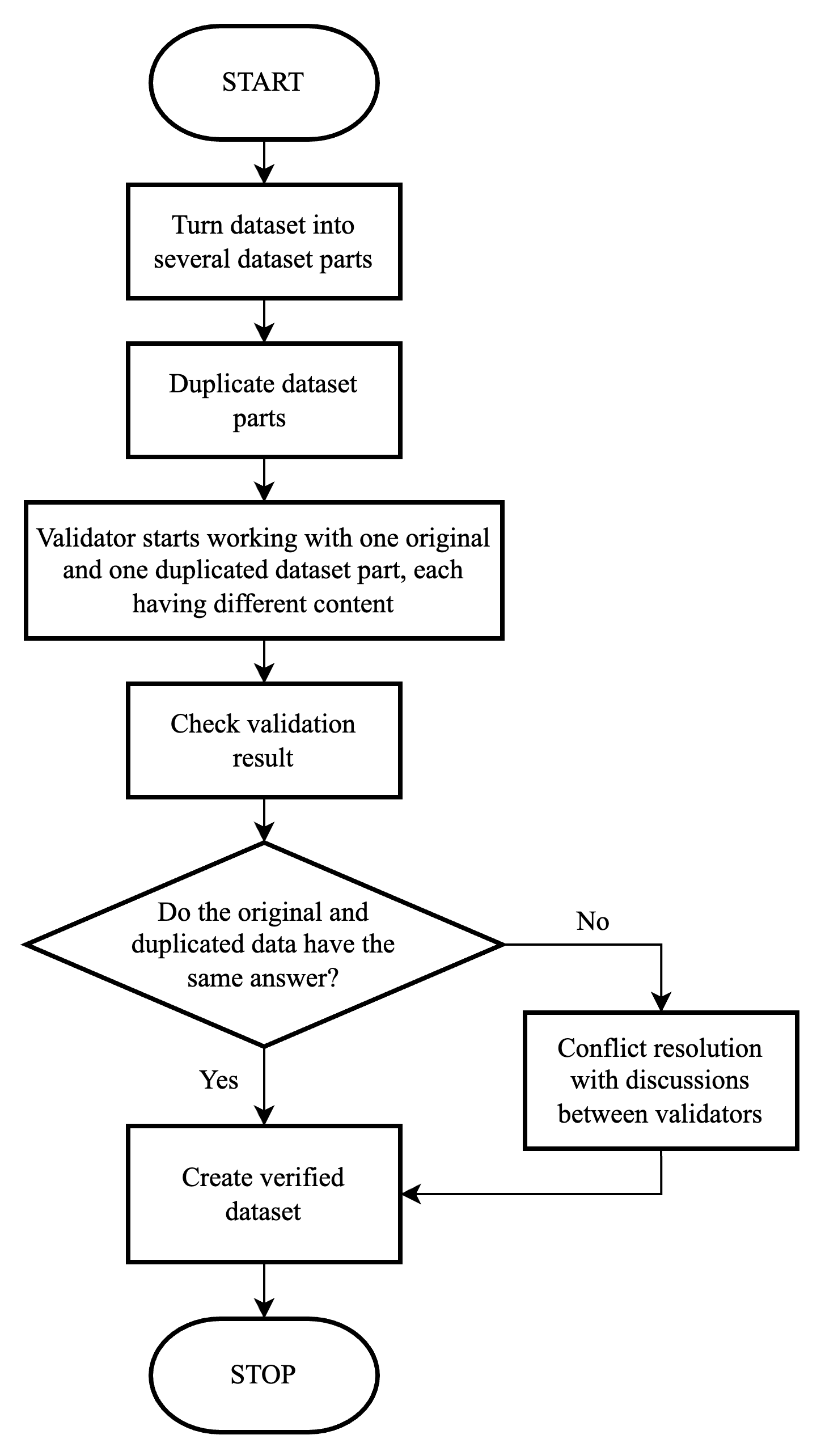}
  \caption{Dataset validation diagram}
  \label{fig:dataset-validation}
\end{figure}

We use some steps in dataset validation including consistency testing of validator performance by doing cross validation as shown in Fig. \ref{fig:dataset-validation}. The validation process begins by dividing 500 context-question-answer pairs into five different dataset parts (A, B, C, D, and E), each part containing the same number of data. These dataset parts are then duplicated (F, G, H, I, and J). Each original has the same content with corresponding duplicate: A with F, B with G, C with H, D with I, and E with J. Each expert validates one original data and one duplicated data with different content. This approach facilitates cross-validation to maintain consistency in the dataset. 

Validation process assisted by five experts, each validator does their task in one week. Once all of the data has been validated by the experts, we proceed to evaluate the data checking the answer similarity. We compare the validation result, fixed small typo by ourselves, 4.2\% of the data have non similar answer, we use discussion mechanisms with related experts to determine the correct answer. The final outcome of this process is a set of 500 context-question-answer pairs that have been validated by the experts.

\subsection{Dataset Content}

The final result of QASiNa dataset is a JSON array file, containing 66 data based on context, example of data is shown on Fig. \ref{fig:data-example}. Each context has attributes \( \texttt{context\_id} \), \( \texttt{context} \), \( \texttt{question\_answers} \), and \( \texttt{context\_length} \). The context length is the number of characters in context, ranges from minimum of 713 to maximum of 1999, with mean of 1629.5 and median of 1689.5. Within each context data, there are several question-answer pairs, which each question-answer pair has attributes \( \texttt{type} \), \( \texttt{question} \), \( \texttt{answer} \), and \( \texttt{answer\_start} \). There are five types of questions: what, when, where, who, and how many, with the distribution of question types shown in Table \ref{table:question-type}.

\begin{table}
\centering
\caption{Number of Data Based on Question Type}
\label{table:question-type}
\centering
\begin{tabular}{>{\centering\arraybackslash}m{2cm} >{\centering\arraybackslash}m{2cm}}
\hline
\textbf{Question Type} & \textbf{Number of Data} \\
\hline
\hline
who & 226 \\
what & 178 \\
how many & 38 \\
where & 38 \\
when & 20 \\
\hline
\textbf{total} & \textbf{500} \\
\hline
\end{tabular}
\end{table}

\section{Evaluation}

We evaluate the dataset using the transfer learning approach with mBERT, XLM-R, and IndoBERT with also test the Chat GPT-3.5 and GPT-4. The evaluation metrics used in this evaluation are Exact Match (EM), F1-Score, and Substring Match evaluation. We choose the Substring Match evaluation to assess the interpretations made by Chat GPT. Substring Match metric assigns a True value if the label is a substring of the generated answer and otherwise is False.

\subsection{Transfer Learning}\label{BERT}

We chose mBERT \cite{devlin2018bert}, XLM-R \cite{conneau2019unsupervised}, and IndoBERT \cite{koto2020indolem} for extractive question answering tasks in Indonesian Language. The tools used are 16 GB Kaggle GPU P100, Huggingface \( \texttt{Trainer} \) with \( \texttt{AutoModelForQuestionAnswering} \) and WandB for monitoring. To fine-tune the selected language models, we use tokenized context and question using each language model tokenizer as input, then the label is answer token position in the context. We use training data from Indonesian translation of SQuAD v2.0 (SQuAD-ID) \cite{muis2020sequence} and conducted grid search hyperparameters tuning, the used parameters are learning rate of 2e-5 and 2e-6, batch sizes of 8 and 16, weight decay of 0.01, and 5 epochs. The best fine-tuned model from each language model is selected by the highest EM score.

We evaluate fine-tuned mBERT, XLM-R, and IndoBERT with SQuAD-ID test set and QASiNa dataset. In addition, we also conducted comparative testing by randomly sampling an equal number of 500 data from the SQuAD-ID test set with \( \texttt{random\_state=0} \), namely 500 SQuAD-ID. The evaluation results of the fine-tuned model are presented in Table \ref{table:eval-lm}.

\begin{table}
\centering
\caption{Evaluation Results using Language Model}
\label{table:eval-lm}
\begin{tblr}{
  width = \linewidth,
  colspec = {Q[190]Q[180]Q[100]Q[100]Q[100]},
  row{1} = {c},
  row{2} = {c},
  cell{1}{1} = {r=2}{},
  cell{1}{2} = {r=2}{},
  cell{1}{3} = {c=3}{},
  cell{3}{1} = {r=3}{},
  cell{3}{3} = {c},
  cell{3}{4} = {c},
  cell{3}{5} = {c},
  cell{4}{3} = {c},
  cell{4}{4} = {c},
  cell{4}{5} = {c},
  cell{5}{3} = {c},
  cell{5}{4} = {c},
  cell{5}{5} = {c},
  cell{6}{1} = {r=3}{},
  cell{6}{3} = {c},
  cell{6}{4} = {c},
  cell{6}{5} = {c},
  cell{7}{3} = {c},
  cell{7}{4} = {c},
  cell{7}{5} = {c},
  cell{8}{3} = {c},
  cell{8}{4} = {c},
  cell{8}{5} = {c},
  cell{9}{1} = {r=3}{},
  cell{9}{3} = {c},
  cell{9}{4} = {c},
  cell{9}{5} = {c},
  cell{10}{3} = {c},
  cell{10}{4} = {c},
  cell{10}{5} = {c},
  cell{11}{3} = {c},
  cell{11}{4} = {c},
  cell{11}{5} = {c},
  vline{2-3} = {1-2,3-5,6-8,9-13}{},
  vline{3} = {4-5,7-8,10-11}{},
  hline{1} = {-}{},
  hline{1,3,6,9,12} = {-}{},
  hline{2} = {3-5}{},
}
\textbf{Model} & \textbf{Dataset} & \textbf{Evaluation Metrics} & & \\
    && \textbf{EM} & \textbf{F1-Score} & {\textbf{Substring}\\\textbf{Match}} \\
\hline 
{IndoBERT\cite{koto2020indolem} \\ \scriptsize indobert-base-p1 (124.5M)} & SQuAD-ID & 39.53 & 59.20 & 54.48 \\
    & 500 SQuAD-ID & 37.00 & 57.65 & 53.40 \\
    & QASiNa & 42.40 & 57.77 & 49.00 \\
{XLM-R\cite{conneau2019unsupervised} \\ \scriptsize xlm-roberta-base (279M)} & SQuAD-ID & \textbf{45.29} & \textbf{64.53} & \textbf{59.51} \\
    & 500 SQuAD-ID & \textbf{42.80} & \textbf{63.87} & \textbf{58.20} \\
    & QASiNa & \textbf{61.20} & \textbf{75.94} & \textbf{70.00} \\
{mBERT\cite{devlin2018bert} \\ \scriptsize bert-base-multilingual-cased (179M)} & SQuAD-ID & 43.62 & 62.33 & 56.97 \\
    & 500 SQuAD-ID & 41.00 & 60.94 & 54.80 \\
    & QASiNa & 58.40 & 71.76 & 64.60
\end{tblr}
\end{table}

The best model selected based on the highest EM score, as in the religious domain, precise answers without any interpretation or excessive information are prioritized. We use scaling from 0 to 100 for the evaluation metrics. 

The XLM-R model outperforms other models for the QASiNa dataset, achieving an EM score of 61.20, which is 2.80 points higher than mBERT and 18.80 points higher than IndoBERT. The F1-score and Substring Match also indicate that the XLM-R model yields the highest score on both. From each language model, we can see that the F1-Score values are lower than the Substring Match values, indicating that interpretation performed by the language model is low. This F1-score and Substring Match comparison will be further analyzed using Chat GPT-3.5 and GPT-4 in Subsection \ref{GPT}.

\begin{table}
\centering
\caption{Evaluation Metrics for Each Question Type using XLM-R}
\label{table:eval-qt}
\begin{tblr}{
  row{1} = {c},
  row{2} = {c},
  cell{1}{1} = {r=2}{},
  cell{1}{2} = {c=3}{},
  cell{3}{2} = {c},
  cell{3}{3} = {c},
  cell{3}{4} = {c},
  cell{4}{2} = {c},
  cell{4}{3} = {c},
  cell{4}{4} = {c},
  cell{5}{2} = {c},
  cell{5}{3} = {c},
  cell{5}{4} = {c},
  cell{6}{2} = {c},
  cell{6}{3} = {c},
  cell{6}{4} = {c},
  cell{7}{2} = {c},
  cell{7}{3} = {c},
  cell{7}{4} = {c},
  vline{2} = {1-2, 3-7}{},
  hline{1,3,8} = {-}{},
  hline{2} = {2-4}{},
}
\textbf{Question Type } & \textbf{Evaluation Metrics } & & \\
    & \textbf{EM} & \textbf{F1-Score} &{\textbf{Substring}\\\textbf{Match}} \\
\hline
who & 65.49 & 78.37 & 73.45 \\
what & 58.43 & 72.52 & 69.10 \\
where & 55.26 & 75.16 & 65.79 \\
how many & 57.89 & 76.38 & 60.53 \\
when & 55.00 & 79.60 & 65.00
\end{tblr}
\end{table}

We use XLM-R to evaluate QASiNa dataset based on available question types. From the experiments in Table \ref{table:eval-qt}, we can see that the EM scores range from the lowest for the "when" question type at 55.00 to the highest for the "who" question type at 65.49. The difference between the highest and lowest values for each evaluation metric is 10.49 for EM, 7.08 for F1-Score, and 12.93 for Substring Match. The difference range of the scores are not excessively high, indicating data for each question type is good.

\subsection{Generative Model with Chat GPT}\label{GPT}

The utilization of ChatGPT continues to grow across a wide range of domains, including the religious domain. One of the objectives of QASiNa is to assess ChatGPT's reasoning capabilities for questions within the context of the Sirah Nabawiyah, in the task of extractive question answering. The outcomes of this evaluation will be compared with the abilities of language models that were evaluated in the Subsection \ref{BERT}. We conduct testing on the ability of Chat GPT using API \( \texttt{gpt-3.5-turbo} \) and \( \texttt{gpt-4} \) to answer questions with given context in extractive way. Our ChatGPT prompt consists of instruction, context, and question, so the Chat GPT will return the answer as given instruction. We use Python 3 to call Chat GPT API with prompt as the following code.

\lstset{basicstyle=\scriptsize}
\begin{lstlisting}
def build_prompt(context, question):
    return f"""{context}

Berdasarkan konteks pada konteks yang diberikan sebelumnya, berikan jawaban dengan tipe ekstraktif tentang pertanyaan berikut. Kata-kata pada jawaban hanya boleh diambil dari konteks yang diberikan. Jawaban dibuat singkat dan tidak boleh ada penjelasan.

{question}
"""
\end{lstlisting}

\begin{table}
\centering
\caption{Chat GPT Evaluation Results}
\label{table:gpt-eval}
\begin{tblr}{
  width = \linewidth,
  colspec = {Q[220]Q[245]Q[100]Q[157]Q[190]},
  row{1} = {c},
  row{2} = {c},
  cell{1}{1} = {r=2}{},
  cell{1}{2} = {r=2}{},
  cell{1}{3} = {c=3}{0.477\linewidth},
  cell{3}{1} = {r=2}{},
  cell{3}{3} = {c},
  cell{3}{4} = {c},
  cell{3}{5} = {c},
  cell{4}{3} = {c},
  cell{4}{4} = {c},
  cell{4}{5} = {c},
  cell{5}{1} = {r=2}{},
  cell{5}{3} = {c},
  cell{5}{4} = {c},
  cell{5}{5} = {c},
  cell{6}{3} = {c},
  cell{6}{4} = {c},
  cell{6}{5} = {c},
  cell{7}{1} = {r=2}{},
  cell{7}{3} = {c},
  cell{7}{4} = {c},
  cell{7}{5} = {c},
  cell{8}{3} = {c},
  cell{8}{4} = {c},
  cell{8}{5} = {c},
  vline{2-3} = {1-8}{},
  vline{3} = {4,6,8}{},
  hline{1,3,5,7,9} = {-}{},
  hline{2} = {3-7}{},
}
\textbf{Model} & \textbf{Dataset} & \textbf{Evaluation Metrics} & & \\
               & & \textbf{EM} & \textbf{F1-Score} & {\textbf{Substring}\\\textbf{Match}} \\
\hline
{XLM-R\\{\scriptsize (279M)}} & 500 SQuAD-ID & \textbf{42.80} & \textbf{63.87} & \textbf{58.20} \\
               & QASiNa & \textbf{61.20} & \textbf{75.94} & \textbf{70.00} \\
{ChatGPT-3.5\\{\scriptsize (gpt-3.5-turbo)}} & 500 SQuAD-ID & 11.60 & 37.86 & 51.00 \\
               & QASiNa & 30.20 & 58.92 & 83.20 \\
{ChatGPT-4\\{\scriptsize (gpt-4)}} & 500 SQuAD-ID & 4.20 & 30.61 & 61.80 \\
               & QASiNa & 4.60 & 37.36 & 93.60
\end{tblr}
\end{table}

Table \ref{table:gpt-eval} is complete results of the evaluation using Chat GPT and the comparison with XLM-R. Due to the limited resources we didn't evaluate using all of the SQuAD-ID dataset, we use sampled 500 SQuAD-ID and QASiNa dataset in this experiment. We can observe that for both Chat GPT-3.5 and GPT-4, the EM and F1-score are lower than Substring Match, which means there are many other words present besides the actual answer. This concludes that providing extractive instructions is not sufficient to make the generative LLM produce extractive answers. Chat GPT intends to do excessive interpretations, while interpretation within the context of religious domains by the AI is prohibited. This experiment lead us to the conclusion that current Chat GPT (gpt-3.5-turbo and gpt-4) is not suitable for finding extractive answers for questions related to religious domain.

\section{Conclusion and Future Works}

We create a new dataset named Question Answering Sirah Nabawiyah (QASiNa) dataset, which employs a previously unexplored domain, Sirah Nabawiyah in the Indonesian language. To evaluate the language models, we conducted testing using the transfer learning approach for language models with mBERT \cite{devlin2018bert}, XLM-R \cite{conneau2019unsupervised}, and IndoBERT \cite{koto2020indolem}. We also evaluate the performance of LLM Chat GPT (gpt-3.5-turbo and gpt-4) \cite{chatgpt} to solve QASiNa dataset. 

By using three language models with data from SQuAD-ID \cite{muis2020sequence}, randomly sampled 500 SQuAD-ID and QASiNa dataset concludes XLM-R as the best model. We also evaluated Chat GPT-3.5 and GPT-4 by providing instruction prompt for extractive answers based on the given context. The conclusion drawn from this evaluation was that the EM and F1-scores of Chat GPT are lower compared to XLM-R, on the other side Chat GPT has high Substring Match score. This experiment concludes that Chat GPT tends to provide excessive interpretations even after providing the context of question.

Research on question answering in the religious domain remains relatively rare, making it an intriguing field, especially in the current LLM's competitions. Conducting QA research in the religious domain highlights the presence of a domain-specific set of issues that needs further attention, especially when the religious domain has strict rule about giving interpretation. This emphasizes the need for technological advancements to ensure the preservation of the values held by religious community. Further studies to advance this research could involve increasing the dataset size and variations of LLM to be analyzed. Furthermore, we can do research about way to make LLM better at answering religious QA, methods like In Context Learning and context-based token filtering are interesting topics. Finally, we intend to make the proposed QASiNa dataset publicly available to the research community.

\section*{Acknowledgment}

The authors thank the Indonesia Endowment Fund for Education (LPDP) for funding this research. We also thank to the domain experts who have assisted in validating the datasets.

\bibliographystyle{IEEEtran}
\bibliography{Bibliography.bib}
\end{document}